\documentclass[lettersize,journal]{IEEEtran}
\usepackage{amsmath,amsfonts}
\usepackage{algorithmic}
\usepackage{algorithm}
\usepackage{listings}
\usepackage{array}
\usepackage[caption=false,font=normalsize,labelfont=sf,textfont=sf]{subfig}
\usepackage{textcomp}
\usepackage{stfloats}
\usepackage{url}
\usepackage{verbatim}
\usepackage{graphicx}
\usepackage{cite}
\usepackage{bm}
\usepackage{booktabs}
\usepackage[table]{xcolor}
\begin{document}

\title{Exploring Fine-Grained Image-Text Alignment for Referring Remote Sensing Image Segmentation}

\author{ {Sen Lei,  Xinyu Xiao, Tianlin Zhang, Heng-Chao Li, Zhenwei Shi, Qing Zhu}

\IEEEcompsocitemizethanks{
\IEEEcompsocthanksitem
This work was supported in part by the National Natural Science Foundation of China under Project 42230102, the National Natural Science Foundation of China under Grant 62271418, Grant 62125102, and Grant U24B20177, in part by the Natural Science Foundation of Sichuan Province under Grant 2023NSFSC0030, and in part by the Fellowship of China National Postdoctoral Program for Innovative Talents (No. BX20240291). (\textit{Corresponding author: Xinyu Xiao})
} 

\thanks {Sen Lei and Heng-Chao Li are with the School of Information Science and Technology, Southwest Jiaotong University, Chengdu 611756, China. (email: senlei@swjtu.edu.cn, lihengchao\_78@163.com)}

\thanks {Xinyu Xiao is with the Company of Ant Group, Hangzhou 688688, China. (email: smilexiao2020@gmail.com)}

\thanks {Tianlin Zhang is with the Luoyang Institute of Electro-optical Equipment, AVIC, Luoyang, 471000, China. (email: zhangtianlin17@mails.ucas.ac.cn)}

\thanks {Zhenwei Shi are with the Image Processing Center, School
of Astronautics, and the State Key Laboratory of Virtual Reality Technology,
and Systems, Beihang University, Beijing 100191, China. (email: shizhenwei@buaa.edu.cn)}

\thanks {Qing Zhu is with the Faculty of Geosciences and Engineering, Southwest Jiaotong University, Chengdu 611756, China. (email: zhuq66@263.net)}

} 

\markboth{Journal of \LaTeX\ Class Files,~Vol.~14, No.~8, August~2021}%
{Shell \MakeLowercase{\textit{et al.}}: A Sample Article Using IEEEtran.cls for IEEE Journals}

\maketitle

\begin{abstract}
Given a language expression, referring remote sensing image segmentation (RRSIS) aims to identify ground objects and assign pixel-wise labels within the imagery. The one of key challenges for this task is to capture discriminative multi-modal features via text-image alignment. However, the existing RRSIS methods use one vanilla and coarse alignment, where the language expression is directly extracted to be fused with the visual features. In this paper, we argue that a ``fine-grained image-text alignment'' can improve the extraction of multi-modal information. To this point, we propose a new referring remote sensing image segmentation method to fully exploit the visual and linguistic representations. Specifically, the original referring expression is regarded as context text, which is further decoupled into the ground object and spatial position texts. The proposed fine-grained image-text alignment module (FIAM) would simultaneously leverage the features of the input image and the corresponding texts, obtaining better discriminative multi-modal representation. Meanwhile, to handle the various scales of ground objects in remote sensing, we introduce a Text-aware Multi-scale Enhancement Module (TMEM) to adaptively perform cross-scale fusion and intersections. We evaluate the effectiveness of the proposed method on two public referring remote sensing datasets including RefSegRS and RRSIS-D, and our method obtains superior performance over several state-of-the-art methods. The code will be publicly available at https://github.com/Shaosifan/FIANet.
\end{abstract}


\section{Introduction}


Referring remote sensing image segmentation (RRSIS) aims to identify the desired ground objects from remote sensing images guided by the corresponding textual description. It can help users to extract specific regions by their particular needs and improve the efficiency for remote sensing analysis \cite{yuan2024rrsis}. RRSIS plays an important role in many tasks such as land use categorization, typical object identification, urban management, and environmental monitoring \cite{yuan2021review}. 


\begin{figure}
  \centering
  \includegraphics[width=\linewidth]{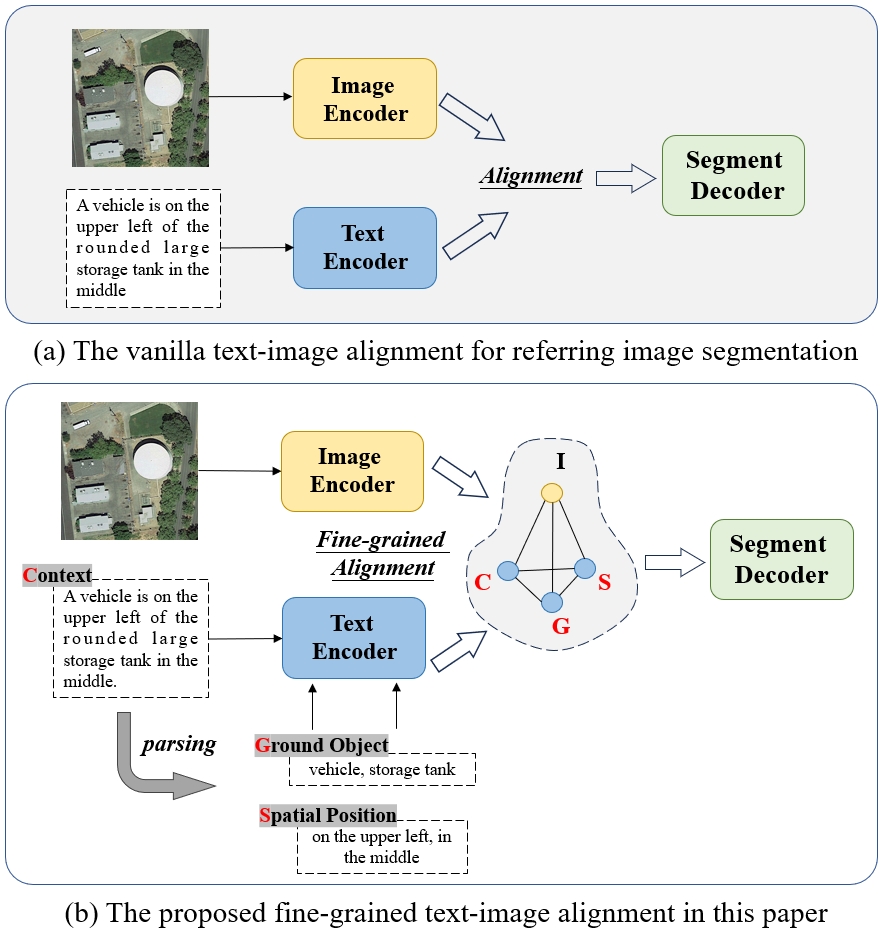}\\
  \caption{ The motivation of the proposed method. (a) shows the vanilla image-text alignment employed in the previous referring image segmentation methods for remote sensing. (b) describes the proposed fine-grained image-text in this article, where the original language expression would be decoupled into ground object fragments and spatial position information. By mining the key elements of images and texts, the association between the image and the referring expression can be clearly constructed, enabling the model to adaptively focus on relevant areas in remote sensing scenarios.     
  }\label{pic:motivation}
\end{figure}

In the past years, deep learning has made great progress in a wide range of remote sensing tasks including super-resolution\cite{lei2017super, lei2021hybrid, lei2021transformer}, scene classification\cite{qiu2024few, tian2024hirenet, qin2024fdgnet}, visual captioning \cite{XiaoWDXP19, XiaoWXP19, XiaoWDXPp19}, object detection \cite{gui2024remote, zhu2024cross, tang2023object}, hyperspectral anomaly detection \cite{wang2023learning, wang2024hyperspectral, zhao2023hyperspectral}, and semantic segmentation\cite{zheng2023farseg++, cao2024bemrf, li2024ida, xie2024landslide}. 
Unlike traditional remote sensing semantic segmentations, RRSIS simultaneously considers the images and textual descriptions and extracts the specific ground objects under text guidance.

There have been several researches in the field of referring image segmentation for natural images over the past few years. 
Early works relied on convolutional neural networks and recurrent neural networks to extract visual and linguistic representations that are subsequently fused by simple concatenation to generate pixel-level results \cite{hu2016segmentation, liu2017recurrent, margffoy2018dynamic}. Then some approaches focus on the elaborate design of image-text alignment to learn more discriminative multi-modal representations \cite{ye2019cross, jing2021locate, hu2020bi, shi2018key}. More recently, Transformer was introduced in the referring image segmentation task and exhibited superior performance than the prior works \cite{yang2022lavt, liu2023multi, kim2022restr}.

Different from natural images, remote sensing imagery usually covers a wide range of ground objects with diverse spatial scales and orientations. It limits these methods designed for natural images to be directly applied in the RRSIS with satisfactory performance \cite{liu2024rotated}. For this point, in the past year, many researchers have focused on the RRSIS task and established two datasets for remote sensing including RefSegRS \cite{yuan2024rrsis} and RRSIS-D \cite{liu2024rotated}, which promotes the development of the field of RRSIS. 
In these methods, both cross-scale and intra-scale information are utilized to accommodate the unique characteristics of remote sensing images, aligning the images with paired texts to achieve a multi-modal representation.


The one of key challenges for this RRSIS task is to learn discriminative multi-modal features via text-image alignment. 
Previous RRSIS methods \cite{yuan2024rrsis,liu2024rotated} typically employed one kind of vanilla and coarse alignment (implicit alignment) of image and text features, as shown in Fig. 1 (a), where the linguistic representation is directly fused with the visual features by leveraging pixel-level attention. This is a concise and direct approach, but \textit{it neglects the intrinsic information within the referring expression and the fine-grained relationship between the image and the textual description}. It might hinder the network from effectively segmentation meeting the complex backgrounds and ground objects with diverse spatial scales in remote sensing.

To handle this issue, we re-examine the vanilla alignment and propose a new paradigm of fine-grained image-text alignment to learn more discriminative multi-modal representations. 
As illustrated in Fig. 1 (b), the original referring sentence is regarded as a \textit{context} expression, and it then is parsed into \textit{ground object} and \textit{spatial position} texts. All these sentence fragments will pass a text encoder and obtain fine-grained linguistic representations.  
By extracting key elements of images and texts, we establish a fine-grained image-text alignment to construct subtle associations between images and their corresponding expressions, enabling the model to adaptively focus on relevant areas in remote sensing scenarios.

In this paper, we propose a novel referring image segmentation method for remote sensing, termed FIANet, from the perspective of fine-grained image-text alignment. Specifically, we design a Fine-grained  Image-text Alignment Module (FIAM) to jointly leverage the features from both input images and the corresponding texts, enabling more discriminative representations across modalities. Meanwhile, to handle the various scales of ground objects in remote sensing, we introduce a Text-aware Multi-scale Enhancement Module (TMEM), which adaptively performs cross-scale fusion and intersections guided by the texts. 
We evaluate the effectiveness of the proposed method on two public referring remote sensing datasets including RefSegRS and RRSIS-D, demonstrating that FIANet achieves superior performance over several state-of-the-art approaches.

The main contributions of this paper are summarized as follows:
\begin{itemize}

\item We propose a novel referring remote sensing image segmentation method named FIANet. Unlike existing methods, FIANet leverages fine-grained image-text alignment to improve multi-modal learning, addressing challenges in handling complex remote sensing scenes. Our method obtains state-of-the-art results on two public remote sensing datasets. 

\item We introduce a fine-grained image-text alignment module to exploit the subtle association between visual and linguistic features, enabling effective segmentations of ground objects under complex backgrounds.  Moreover, we design a text-aware multi-scale enhancement module to leverage cross-scale multi-modal interactions, which can improve FIANet's ability to adapt to ground objects with varying and diverse scales. Comprehensive ablation experiments verify the effectiveness of these designs.

\end{itemize}

The rest parts of this paper are organized as follows. We give a brief description of the background and related work of the referring remote sensing image segmentation in Section  \ref{sec:background}. In Section \ref{sec:method}, we carefully describe our method and the proposed improvements. Many comparative experiments on two public remote sensing datasets and ablation studies are presented in Section \ref{sec:exp}. Finally, conclusions and future works are drawn in Section \ref{sec:conclusion}.

\section{Background and Related Work} \label{sec:background}

\subsection{Referring Image Segmentation for Natural Images}

Referring image segmentation aims to segment a specific target object within an image based on a corresponding textual description, representing a typical multimodal task that has attracted increasing attention.
The pioneering work \cite{hu2016segmentation} utilizes a convolutional neural network and recurrent LSTM to capture visual and linguistic representation. Liu $et~al.$ \cite{liu2017recurrent} proposed a recurrent multimodal interaction model that consists of sequential LSTMs to fulfill word-to-image interaction. Edgar $et~al.$ \cite{margffoy2018dynamic} designed a modular neural network that divides the problem of referring image segmentation into many sub-tasks. These methods fuse visual and linguistic representation by simple concatenation to predict pixel-wise segmentation output, which constrains the capability of joint learning of images and languages. The subsequent works \cite{ye2019cross, jing2021locate, hu2020bi, shi2018key} mainly focus on the elaborate design of image-text alignment to learn more discriminative multi-modal representations. Ye $et~al.$ \cite{ye2019cross} introduced a cross-modal self-attention module to learn the long-range relationship between the visual and linguistic features, as well as a gated multi-level fusion module to integrate multi-level self-attentive features. Jing $et~al.$ \cite{jing2021locate} leveraged a cross-model interaction module on the multi-modal features by the explicit model of position prior.

Recently, Transformer has exhibited superior performance in the referring image segmentation task \cite{yang2022lavt, liu2023multi, kim2022restr}. LAVT \cite{yang2022lavt} employs a vision Transformer \cite{liu2021swin} as the visual encoder and utilizes an early fusion paradigm to perform hierarchical language-aware visual encoding for capturing multi-modal context. Liu $et~al.$ \cite{liu2023multi} designed multi-model mutual attention to better fuse multi-modal information, where the features of inputs are extracted by Swin Transformer and BERT \cite{vaswani2017attention}, respectively. 
However, different from natural images, remote sensing imagery usually covers a wide range of ground objects with diverse spatial scales and orientations, which limits the performance of these methods to generate satisfactory segmentation results.

\subsection{Remote Sensing Referring Image Segmentation and Visual Grounding}

In the past year, researchers have begun to pay attention to the field of Referring Remote Sensing Image Segmentation (RRSIS), and two datasets including RefSegRS \cite{yuan2024rrsis} and RRSIS-D \cite{liu2024rotated} were proposed successively. Yuan $et~al.$ \cite{yuan2024rrsis} tried the first attempt to handle the RRSIS task and proposed a novel Language-Guided Cross-scale Enhancement (LGCE) module to improve the results on small ground objects. Liu $et~al.$ \cite{liu2024rotated} introduced a Rotated Multi-Scale Interaction Network (RMSIN) to mitigate the issues caused by diverse spatial scales and orientations in the remote sensing imagery, in which intra-scale and cross-scale interactions are fully excavated. 

Remote Sensing Visual Grounding (RSVG) aims to localize ground objects with bounding boxes referring to the given textual descriptions. Similar to the RRSIS, the one of key challenges for RSVG is to effectively fuse visual and linguistic representations to predict the object's location. Sun $et~al.$ \cite{sun2022visual} established a new visual ground benchmark dataset for remote sensing and proposed a new model composed of image/language encoders and the corresponding fusion module. Furthermore, Zhan $et~al.$ \cite{zhan2023rsvg} introduced a transformer-based method with multi-level cross-modal feature learning to handle large-scale variations and cluttered backgrounds.
More recently, Kuckreja $et~al.$ \cite{kuckreja2024geochat} proposed a novel grounded large vision-language model that offered multi-task capacity for high-resolution remote sensing images. This work can handle multiple tasks simultaneously, including visual grounding, image/region caption, scene classification, etc.

\section{Methodology}\label{sec:method}


\subsection{Overview of the Proposed Method}

\begin{figure*}
  \centering
  \includegraphics[width=\linewidth]{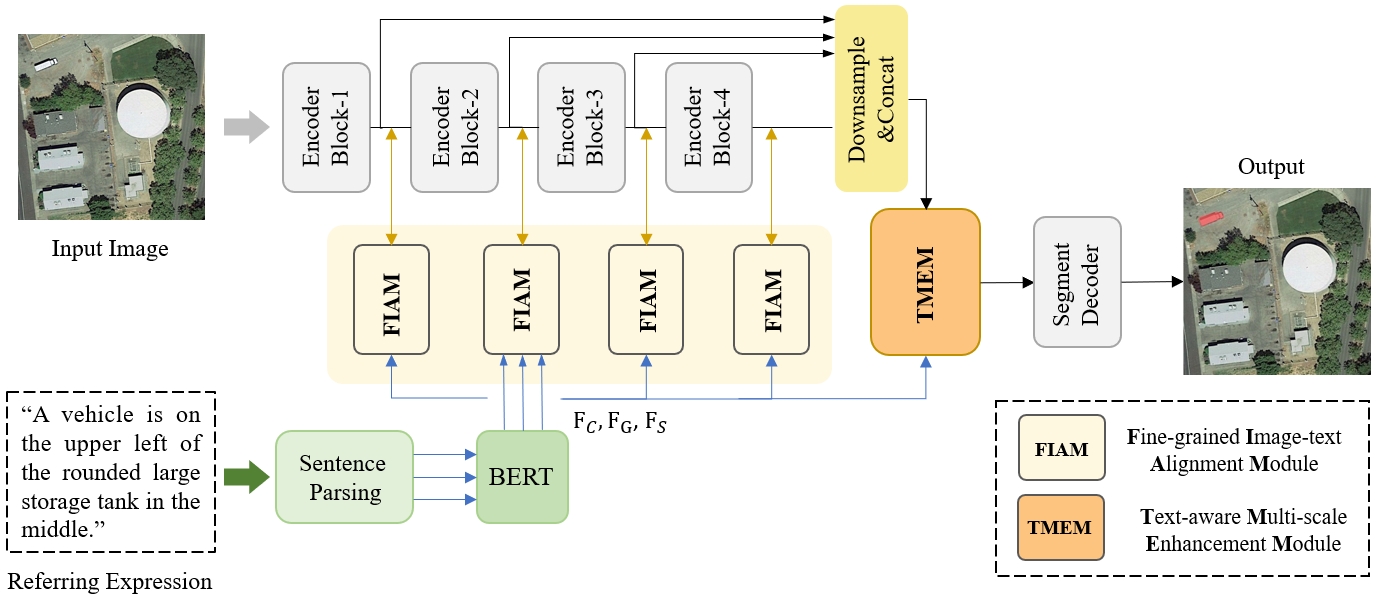}\\
  \caption{The framework of the proposed method. The original textual description is regarded as context expression and further is parsed into two fragments about ground objects and spatial positions. There would be three linguistic features in total, including $F_C$, $F_G$, and $F_S$ which denote the representations extracted by the pre-trained BERT from the original context expression, ground objects, and spatial positions. 
  Fine-grained image-text alignment modules (Sec. 3.2) would subtly align visual and linguistic representations, and the text-aware multi-scale enhancement module (Sec. 3.3) is designed to fuse multi-model representations from different levels. 
  }\label{pic:framework}
\end{figure*}

In this paper, we propose a novel referring remote sensing image segmentation method named FIANet, which is illustrated in Fig. \ref{pic:framework}.   
Similar to the previous works \cite{yuan2024rrsis, liu2024rotated}, the pipeline of FIANet is divided into four procedures: feature extraction, image-text alignment, multi-scale fusion, and segment decoding. 
Visual and linguistic representations are first extracted from the image and its paired referring expression by an image encoder and a text encoder, respectively. Notably, the original textual description is treated as a contextual expression, which we further decompose into two components: one describing ground objects and the other detailing spatial positions. Thus, three linguistic features are obtained, representing the original contextual expression, ground objects, and spatial positions.
Specifically, we employ the Natural Language Toolkit (NLTK) \cite{loper2002nltk} to parse the referring expression based on each dataset's predefined ground object categories. The entire parsing process is conducted offline before training or inference, making it highly efficient. These three linguistic features are extracted by using a pre-trained BERT \cite{devlin2018bert}. 

The hierarchical visual features extracted from various stages of the encoder are subsequently aligned with the corresponding linguistic features, thereby enabling the capture of discriminative multi-modal representations.
For this point, we propose a Fine-grained Image-text Alignment Module (FIAM) to subtly align visual and linguistic representations.
After that, the Text-aware Multi-scale Enhancement Module (TMEM) is implemented to combine these multi-modal representations from different levels, which improves the ability of FIANet to adapt to ground objects with varying and diverse scales
Finally, the enhanced multi-scale representations would be integrated to generate the pixel-wise segmentation by the segment decoder. 

Algorithm \ref{alg:code} provides the pseudocode of FIANet in a PyTorch-like style, where the main components of forward-pass are involved. More details about the FIAM and TMEM will be carefully described in the following subsections.

\begin{algorithm}[t]
\caption{ Pseudocode of FIANet in a PyTorch-like style.}
\label{alg:code}
\definecolor{codeblue}{rgb}{0.25,0.5,0.5}
\lstset{
  backgroundcolor=\color{white},
  basicstyle=\fontsize{7.2pt}{7.2pt}\ttfamily\selectfont,
  columns=fullflexible,
  breaklines=true,
  captionpos=b,
  commentstyle=\fontsize{7.2pt}{7.2pt}\color{codeblue},
  keywordstyle=\fontsize{7.2pt}{7.2pt},
}
\begin{lstlisting}[language=python]
# I, T: input image and the corresponding referring text
# FIAM: fine-grained image-text alignment module
# TMEM: text-aware multi-scale enhancement module
# Out:  referring segmentation result

# parse the text and extract linguisic features
T_C, T_G, T_S = Sentence_Parsing(T)
F_C, F_G, F_S = BERT(T_C, T_G, T_S)

# visual representation and fine-grain alignment
F_I_0 = I
for i in (1, 2, 3, 4) # the encoder has four blocks
    F_I_i = Encoder_Block_i(F_I_i-1)
    F_I_i = FIAM(F_I_i, F_C, F_G, F_S)

# multi-scale enhancement with visual/linguisic features
F_I_1, F_I_2, F_I_3 = Downsample(F_I_1, F_I_2, F_I_3)
F_cat = Concat(F_I_1, F_I_2, F_I_3, F_I_4)
F_cat = TMEM(F_cat, F_C)

# obtain final result 
Out = Segmenat_Decoder(F_cat)

\end{lstlisting}
\end{algorithm}


\subsection{Fine-Grained Image-Text Alignment}

\begin{figure}
  \centering
  \includegraphics[width=\linewidth]{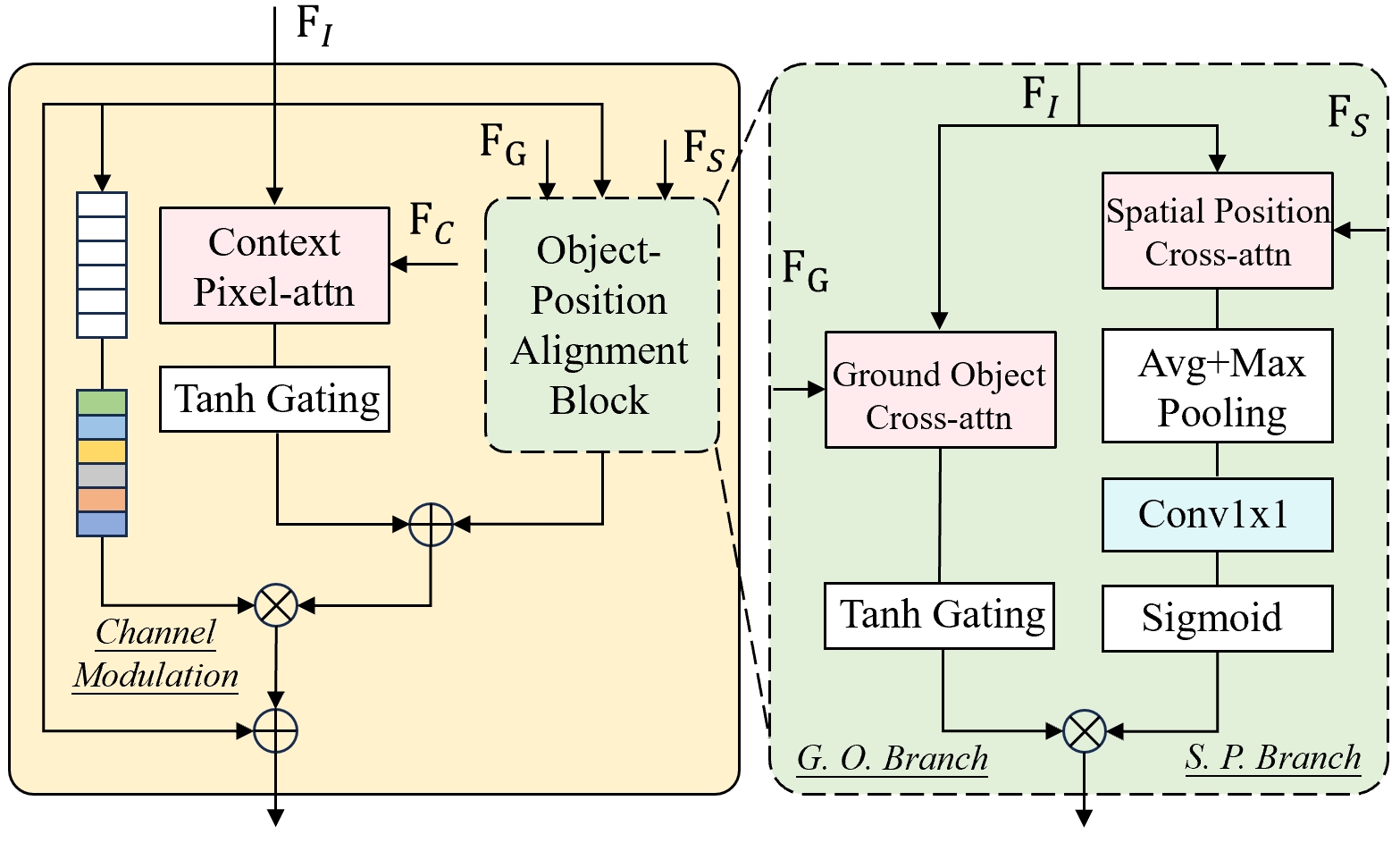}\\
  \caption{ The illustration of the Fine-grained Image-text Alignment Module (FIAM) which aims to obtain discriminative multi-modal representation using visual and fine-grained linguistic features.  
  }\label{pic:FIAM}
\end{figure}

Different from the traditional image-text alignment in the previous methods, we introduce a new multi-modal fusion manner from the perspective of fine-grained alignment to capture more discriminative representations. Concretely, given the visual feature $\bm{F}_{I} \in\mathbb{R}^{C \times H \times W}$, and the linguistic features $\bm{F}_{C} \in\mathbb{R}^{N_C \times D}$, $\bm{F}_{G} \in\mathbb{R}^{N_G \times D}$,  and $\bm{F}_{S} \in\mathbb{R}^{N_S \times D}$, the Fine-grained Image-text Alignment Module (FIAM) is introduced to perform deep intersections between these visual and linguistic features. Here, $C$, $H$ and $W$ denote the number of channels, height, and width of the visual feature maps. Moreover, $D$ is the dimension of word embeddings, and $N_C$, $N_G$, and $N_S$ represent the length of context, ground object, and spatial position expressions.  
The detailed structure of FIAM is shown in Fig. \ref{pic:FIAM}. The core components of the FIAM are the object-position alignment block, context alignment, and channel modulation, which are carefully described below.

\textit{1) Object-Position Alignment Block:} For each FIAM, we propose an Object-Position Alignment Block (OPAB) to perform the deep intersection of features from the ground object and spatial position with the visual representation. 
This block enables precise alignment of object-related and spatial features, and allows the model to capture more accurate relationships between objects and their positions within images, thereby enhancing referring segmentation performance.
The detailed structure of OPAB is illustrated in the Fig. \ref{pic:FIAM}. Specifically, a dual-branch structure is constructed by a \textit{Ground Object Branch} and a \textit{Spatial Position Branch}. The object branch is established to directly perform multi-fusion between the textual features of ground objects and the visual features, which can enhance the discriminative ability of the model on the referent target. The main part of the ground object branch is a ground object cross-attention block that can integrate the visual feature $\bm{F}_{I}$ and the textual feature $\bm{F}_{G}$. Here, we take the $\bm{F}_{I}$ as the query, and the $\bm{F}_{G}$ as the key and value to achieve feature fusion. 
This implementation can be defined as:
\begin{equation}
\bm{F}_{IG} = {\rm Softmax}(\frac{\bm{F}_I \bm{W}_q^{ig} \cdot \bm{F}_G (\bm{W}_k^{ig})^T}{\sqrt{C}}) \cdot \bm {F}_G \bm{W}_v^{ig},
\end{equation}
where $\bm{W}_q^{ig}$, $\bm{W}_k^{ig}$, and $\bm{W}_v^{ig}$ are the linear projection matrices which are responding to the query, key, and value, respectively. The $C$ is the dimension of the query.

Moreover, the image-language feature $\bm{F}_{IG}$ is further modulated by a tanh gate to provide more local details and produce the output $\bm{F}_{GOB}$ for this ground object branch. The calculation can be defined as follows
\begin{equation}
   \bm{F}_{GOB} = {\rm Tanh\_Gate}(\bm{F}_{IG}) \cdot \bm{F}_{IG}, 
\end{equation}
where the $\rm Tanh\_Gate(\cdot)$ denotes a series of operations, sequentially including linear projection, ReLU activation, linear projection, and Tanh activation. 

The spatial position branch is designed to capture the spatial prior guided by the original visual feature $\bm{F}_{I}$ and the textual features of positional description $\bm{F}_{S}$. Concretely, the $\bm{F}_{I}$ and $\bm{F}_{S}$ will go through a cross-attention, where $\bm{F}_{I}$ is taken as the query and $\bm{F}_{G}$ is the key and value: 
\begin{equation}
\bm{F}_{IS} = {\rm Softmax}(\frac{\bm{F}_I \bm{W}_q^{IS} \cdot \bm{F}_S (\bm{W}_k^{IS})^T}{\sqrt{C}}) \cdot \bm{F}_S \bm{W}_v^{IS}.
\end{equation}
Then the $\bm{F}_{IS}$ is input into a series of layers to generate spatial attention, where average and maximum pooling, $1 \times 1$ convolution, and sigmoid nonlinearity are implemented. It is shown in Fig. \ref{pic:FIAM} and mathematically described as follows
\begin{equation}
\begin{aligned}
  &\bm{F}_{cat} = {\rm Concat}( {\rm Avg\_Pool}(\bm{F}_{IS}), \rm Max\_Pool(\bm{F}_{\textit{IS}})), \\ 
  &\bm{F}_{SPB} = {\rm Sigmoid}({\rm Conv}(\bm{F}_{cat})), \\
\end{aligned}    
\end{equation}
where the $\bm{F}_{SPB}$ is the output of this spatial position branch and is regarded as one kind of spatial prior, which involves the multi-modal information of visual and textual features. The $\bm{F}_{SPB}$ is further integrated with ground object features to acquire the final output of OPAB:

\begin{equation}
\bm{F}_{OPAB} = \bm{F}_{GOB} \otimes \bm{F}_{SPB},
\end{equation}
where $\otimes$ denotes element-wise multiplication.  
The output of OPAB is the fine-alignment features that simultaneously consider the ground object and the corresponding spatial attention, and can help the FIAM to obtain more discriminative representation referring to the specific objects. 

\textit{2) Context Alignment with Visual Features:}
Apart from the object-position alignment, we also introduce context alignment to capture the global relationships between image and linguistic features. The original referring expression is treated as a contextual description, which contains more contextual information compared to the sentence fragments of the ground object and spatial position. Given the linguistic contextual feature $\bm{F}_C$ and the visual feature $\bm{F}_I$, one pixel-attention is employed to combine these two features:
\begin{equation}
\bm{F}_{IC} = {\rm Pixel\_Attention} (\bm{F}_{I}, \bm{F}_{C}),
\end{equation}
Here, the pixel attention is implemented by the Pixel-Word Attention Module (PWAM)\cite{yang2022lavt}, which aligns the visual representations with the language features of the original description. Similar to the ground object cross-attention and spatial position cross-attention, we use the $\bm{F}_{I}$ as the query and the $\bm{F}_{C}$ as the key and value in the pixel attention. Moreover, the image-language feature $\bm{F}_{IC}$ is also modulated by a tanh gate. The calculation can be defined as follows
\begin{equation}
   \hat{\bm{F}}_{IC} = {\rm Tanh\_Gate}(\bm{F}_{IC}) \cdot \bm{F}_{IC}, 
\end{equation}
After acquiring the $\hat{\bm{F}}_{IC}$ and the $\bm{F}_{OPAB}$, the multi-modal features $\bm{F}_{CGS}$ further be obtained by combining these two:
\begin{equation}
\bm{F}_{IO} = \hat{\bm{F}}_{IC} + \bm{F}_{OPAB},
\end{equation}

\textit{3) Channel Modulation:} In order to encourage information exchange across channels, we here proposed a channel modulation operator to readjust the extracted multi-modal features, which can further enhance the discriminative ability of the proposed method. Specifically, channel-wise dependencies can be obtained by 
\begin{equation}
\bm{c} = \sigma(\bm{W}_2\delta(\bm{W}_1\cdot{\rm Avg\_{Pool}}(\bm{F}_{IO}))), 
\end{equation}
where the $\bm{W}_1$ and $\bm{W}_2$ are learned weights to perform channel shrink and channel expansion, respectively. The $\delta$ denotes the ReLU function and the $\sigma$ indicates the sigmoid function.    

Then the channel-weight $\bm{c}$ would be utilized to recalibrate the multi-modal feature $F_{IO}$ to acquire the final output of the FIAM with the original input $\bm{F}_{I}$. The calculation is as follows:  
\begin{equation}
\bm{F}_{FIAM} = \bm{c} \otimes \bm{F}_{IO} + \bm{F}_{I},
\end{equation}

Overall, through fine-grained image-text alignment, the $\bm{F}_{FIAM}$ effectively integrates the visual feature with text features at different levels covering the context, ground objects, and spatial positions. Compared to existing methods, the proposed network can acquire more fine-grained informative features, thereby enabling more accurate pixel-level segmentation results.


\subsection{Text-Aware Multi-Scale Enhancement}

\begin{figure}
  \centering
  \includegraphics[width=\linewidth]{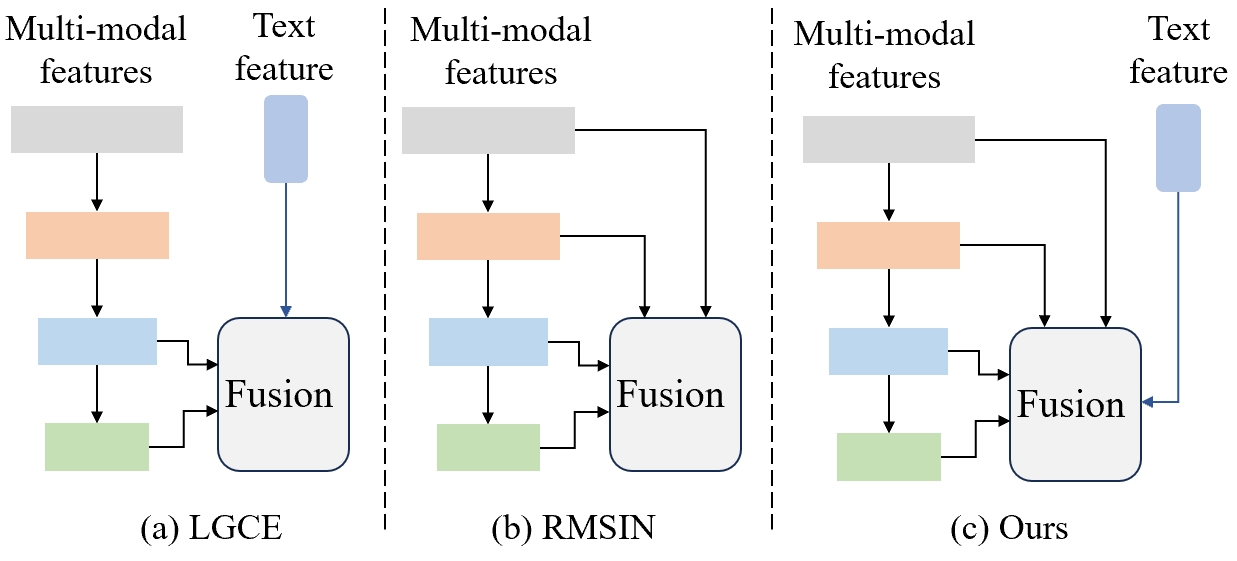}\\
  \caption{ The comparisons of cross-scale interaction within LGCE \cite{yuan2024rrsis}, RMSIN \cite{liu2024rotated}, and our proposed method. Different from these two works, our method can fully explore the multi-scale information of visual representations with text features.  
  }\label{pic:scale_fusion}
\end{figure}

\begin{figure}
  \centering
  \includegraphics[width=7cm]{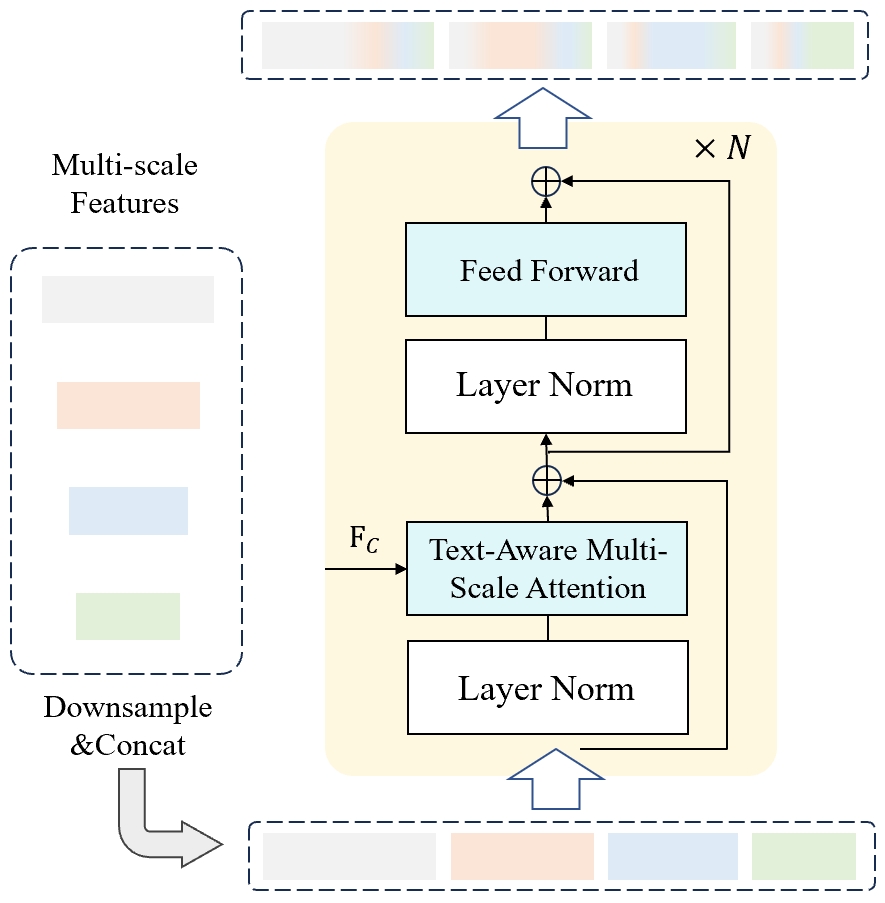}\\
  \caption{ The illustration of Text-Aware Multi-Scale Enhancement Module (TMEM). Before input into the TMEM, the multi-scale features need to be downsampled and concatenated.
  }\label{pic:TMEM}
\end{figure}

The ground objects in remote sensing images exhibit a wide of scales, hindering the effective extraction of referring objects.
Therefore, cross-scale interaction, which leverages features from different scales, plays an important role in this RRSIS task.
As shown in Fig. \ref{pic:scale_fusion} (a) and Fig. \ref{pic:scale_fusion} (b), LCGE \cite{yuan2024rrsis} explores the cross-scale correlation with only two scales assisted by text guidance, and RMSIN \cite{liu2024rotated} employs cross-scale interaction between all the four scales' features without text guidance. These methods have not fully explored the multi-scale information of visual representations with linguistic features. Drawing inspiration from these methods, we propose a Text-Aware Multi-Scale Enhancement Module (TMEM) to effectively leverage multi-scale visual and linguistic features, where the schematic diagram is shown in Fig. \ref{pic:scale_fusion} (c). We compare the proposed TMEM with the cross-scale correlation approaches designed in LCGE and RMSIN and verify the superior performance of the TMEM over these two approaches. More detailed information can be found in the next section (Sec. 4.3).    

Before inputting into the TMEM, all the multi-modal representations from different stages of the image encoder are first preprocessed to ensure they have the same spatial dimensions.
Supposing that $F_I^i$ denotes the output of stage $i$, these features are downsampled with average pooling to the same size and then concatenated along the channel dimension:
\begin{equation}
\begin{aligned}
  &\hat {\bm{F}}_{I}^i = {\rm Downsample}(\bm{F}_I^i), \\ 
  &\hat {\bm{F}}_{cat} = {\rm Concat} (\hat {\bm{F}}_{I}^1, ..., \hat {\bm{F}}_{I}^4), \\
\end{aligned}    
\end{equation}
where the $\hat {\bm{F}}_{I}^i$ denotes the downsampled features of $\bm{F}_I^i$ and the $\hat {\bm{F}}_{cat}$ denotes the concatenated features. 

The next question is how to construct a feasible structure of TMEM to achieve the multi-scale feature fusion. For this point, we design a concise and effective structure based on transformer decoders to capture long-term dependencies across different scales. 
Specifically, the $\hat {\bm{F}}_{cat}$ and linguistic feature $\bm{F}_C$ are fed into TMEM to perform multi-scale fusion. The text-aware multi-scale attention in the TMEM is one kind of multiheaded self-attention (MSA) to perform the deep fusion and intersection with multi-modal features and text guidance. The overall calculation process is as follows:

\begin{equation}
\begin{aligned}
& \bm{z}_0=\hat {\bm{F}}_{cat}, \\
& \bm{z}_i^{'}={\rm Attention}({\rm LN}(\bm{z}_{i-1}), \bm{F}_C) + \bm{z}_{i-1},i=1,…,L_N \\
& \bm{z}_i={\rm MLP}({\rm LN}(\bm{z}_i^{'}))+ \bm{z}_i^{'}, i=1,…,L_N \\
\end{aligned}
\end{equation}
where the $\rm LN$ denotes layer normalization \cite{ba2016layer} and the $\rm MLP$ denotes the multi-layer perceptron which has two layers with GELU nonlinear function \cite{hendrycks2016gaussian}. The $\rm Attention(\cdot)$ represents the text-aware multi-scale attention in the TMEM, and the text representation is integrated into the multi-scale fusion to enhance the discriminative ability for referring objects with diverse scales, which can be computed as:
\begin{equation}
\hat {\bm{z}}_{i-1} = {\rm Softmax}(\frac{{\rm LN}(\bm{z}_{i-1}) \bm{W}_q^{i-1} \cdot \bm{F}_C (\bm{W}_k^{i-1})^T}{\sqrt{C^{'}}}) \cdot \bm{F}_C \bm{W}_v^{i-1}.
\end{equation}
  
After obtaining the multi-scale enhanced features, the output of TMEM is split along the channel dimension and is upsampled to the original spatial dimension. These enhanced multi-scaled features are passed through a scale-aware gate \cite{liu2024rotated} and a segment decoder to make the final mask prediction.   


\subsection{Implementation Details}

In this paper, the proposed method is implemented using Pytorch \cite{paszke2019pytorch}. Following the setting of \cite{yuan2024rrsis} and \cite{liu2024rotated}, We utilize the Swin Transformer as the visual backbone, which is pre-trained on ImageNet22K \cite{deng2009imagenet}, and use the BERT from HuggingFace’s Transformer library \cite{wolf2020transformers} as the text encoder. The image encoder and text encoder will be fine-tuned on the remote sensing dataset. 
Referring to the work \cite{liu2024rotated}, we use the combination of cross-entropy loss and dice loss to train our model, where the weight of dice loss is set to 0.1. 

There are two RRSIS datasets including RefSegRS \cite{yuan2024rrsis} and RRSIS-D \cite{liu2024rotated}, and all the images are resized at $480 \times 480$ pixels.
For the RefSegRS and RRSIS-D datasets, we train the model for 60 epochs and 40 epochs, with a learning rate of 5e-5 and 3e-5, respectively. In the training phase, AdamW \cite{loshchilov2017decoupled} is adopted to optimize the model, and weight decay is set to 0.1. 
All the experiments are conducted on an NVIDIA GeForce RTX 4090 GPU with a batch size of 8.  

\section{Experiments}\label{sec:exp}

\begin{table*}[]
\centering
	 \caption{The results of referring image segmentation with different methods on the \textbf{RefSegRS Dataset}. The best Performance is Bold.}
 \resizebox{15cm}{!}{
\begin{tabular}{c|c|ccccc|cc}
\toprule
Methods  & Publication & Pr@0.5  & Pr@0.6  & Pr@0.7  & Pr@0.8   & Pr@0.9  & oIoU   & mIoU            \\  \midrule
LSTM-CNN \cite{hu2016segmentation} & ECCV'2016   & 15.69  & 10.57   & 5.17  & 1.10  & 0.28  & 53.83  & 24.76 \\ 
ConvLSTM \cite{li2018referring} & CVPR'2018   & 31.21  & 23.39   & 15.30  & 7.59   & 1.10  & 66.12  & 43.34  \\ 
CMSA  \cite{ye2019cross}    & CVPR'2019      & 28.07          & 20.25          & 12.71          & 5.61          & 0.83         & 64.53          & 41.47          \\ 
BRINet \cite{hu2020bi}  & CVPR'2020      & 22.56          & 15.74          & 9.85          & 3.52          & 0.50         & 60.16          & 32.87          \\ 
LAVT   \cite{yang2022lavt}  & CVPR'2022  &70.23 &55.53 &30.05 &14.42 &4.07 &76.21 & 57.30          \\ 
CrossVLT \cite{cho2023cross} & TMM'2023  &71.16 &58.28 &34.51 &16.35 &5.06 & 77.44 & 58.84  \\
RMISN \cite{liu2024rotated} & CVPR'2024 & 71.60 &55.97 &31.87 &11.72 &1.93 &71.73 & 57.78  \\  
LGCE \cite{yuan2024rrsis}  & TGRS'2024   &76.55  &67.03 &44.85 &19.04  &5.67  &77.62 &61.90  \\  
\rowcolor{gray!25} FIANet (ours)  & ---  &\textbf{84.09} &\textbf{77.05} &\textbf{61.86} &\textbf{33.41} &\textbf{7.10} &\textbf{78.32} & \textbf{68.67}           \\ \bottomrule
\end{tabular}
} \label{tab:refsegrs_overall}
\end{table*}

\begin{table}[]
\centering
	 \caption{The results on each category of \textbf{RefSegRS Dataset}. The best Performance is Bold.}
 \resizebox{\linewidth}{!}{
\begin{tabular}{c|cccc}
\toprule
category    & LAVT   & RMSIN  & LGCE  & Ours \\  \midrule
road                 &70.33  &66.67   &74.03  &\textbf{74.13}  \\
vehicle              &57.02  &58.66   &63.05  &\textbf{70.15}  \\
car                  &55.13  &57.63   &60.79  &\textbf{68.55}  \\
van                  &38.55  &47.09   &41.60  &\textbf{61.06}  \\
building             &81.92  &76.84   &\textbf{81.99}  &81.34  \\
truck                &53.07  &51.92   &62.69  &\textbf{74.48}  \\
trailer              &44.56  &61.65   &49.82  &\textbf{74.92}  \\
bus                  &52.93  &60.20   &45.36  &\textbf{72.40}  \\
road marking         &5.74   &18.60   &6.66   &\textbf{22.85}  \\
bikeway              &50.26  &50.35   &54.23  &\textbf{61.16}  \\
sidewalk             &57.35  &49.12   &61.68  &\textbf{62.90}  \\
tree                 &57.01  &49.82   &67.68  &\textbf{83.75}  \\
low vegetation       &41.08  &43.73   &43.68  &\textbf{44.84}  \\
impervious surface   &81.51  &76.55   &\textbf{83.18}  &81.53  \\ \midrule
average              &53.32  &54.92   &56.89  &\textbf{66.72}  \\  \bottomrule

\end{tabular}
} \label{tab:refsegrs_class}
\end{table}

\begin{table*}[]
\centering
	 \caption{The results of referring image segmentation with different methods on the \textbf{RRSIS-D Dataset}. The best Performance is Bold.}
 \resizebox{15cm}{!}{
\begin{tabular}{c|c|ccccc|cc}
\toprule
Methods  & Publication & Pr@0.5  & Pr@0.6  & Pr@0.7  & Pr@0.8   & Pr@0.9  & oIoU   & mIoU            \\  \midrule
RRN \cite{li2018RRN} & CVPR'2018   & 51.07  & 42.11   & 32.77  & 21.57  & 6.37  & 66.43  & 45.64 \\
CMSA \cite{ye2019cross} & CVPR'2019   & 55.32  & 46.45  & 37.43  & 25.39   & 8.15  & 69.39  & 48.54  \\ 

LSCM  \cite{hui2020linguistic}    & ECCV'2020      & 56.02          & 46.25          & 37.70          & 25.28          & 8.27        & 69.05          & 49.92          \\ 
CMPC \cite{huang2020referring}  & CVPR'2020      & 55.83          & 47.40          & 36.94          & 25.45          & 9.19        & 69.22         & 49.24          \\ 
BRINet   \cite{hu2020bi}  & CVPR'2020      & 56.90          & 48.77          & 39.12          & 27.03          & 8.73          & 69.88          & 49.65          \\ 
CMPC+   \cite{liu2021cross}  & TPAMI'2021      & 57.65          & 47.51          & 36.97          & 24.33          & 7.78         & 68.64          & 50.24          \\ 
LAVT   \cite{yang2022lavt}  & CVPR'2022  &66.93 &60.99 &51.71 &39.79 &23.99 &76.58 &59.05 \\ 
CrossVLT \cite{cho2023cross} & TMM'2023   &70.38 &63.83 &52.86 &42.11 &\textbf{25.02} & 76.32 & 61.00 \\
LGCE \cite{yuan2024rrsis}     & TGRS'2024  &69.41 &63.06 & 53.46 &41.22 &24.27 &76.24 &61.02   \\  

RMISN \cite{liu2024rotated}  & CVPR'2024 &71.96 &65.76 &55.16 &42.03 &\textbf{25.02} &76.50 &62.27   \\  

\rowcolor{gray!25} FIANet (ours)  & --- &\textbf{74.46} &\textbf{66.96} &\textbf{56.31} &\textbf{42.83} &24.13 &\textbf{76.91} & \textbf{64.01}      \\ \bottomrule
\end{tabular}
} \label{tab:rrsisd_overall}
\end{table*}

\begin{table}[]
\centering
	 \caption{The results on each category of \textbf{RRSIS-D Dataset}. The best Performance is Bold.}
 \resizebox{\linewidth}{!}{
\begin{tabular}{c|cccc}
\toprule
category    & LAVT   & RMSIN  & LGCE  & Ours \\  \midrule
airport                       &66.44  &68.08  &68.11  &\textbf{68.66}  \\
golf field                    &56.53  &56.11  &56.43  &\textbf{57.07}  \\
expressway service area       &76.08  &76.68  &77.19  &\textbf{77.35}  \\
baseball field                &68.56  &66.93  &\textbf{70.93}  &70.44  \\
stadium                       &81.77  &83.09  &\textbf{84.90}  &84.87  \\
ground track field            &81.84  &81.91  &\textbf{82.54}  &82.00  \\
storage tank                  &71.33  &73.65  &73.33  &\textbf{76.99}  \\
basketball court              &70.71  &72.26  &74.37  &\textbf{74.86}  \\
chimney                       &65.54  &68.42  &\textbf{68.44}  &68.41  \\
tennis court                  &74.98  &76.68  &75.63  &\textbf{78.48}  \\
overpass                      &66.17  &\textbf{70.14}  &67.67  &70.01  \\
train station                 &57.02  &\textbf{62.67}  &58.19  &61.30  \\
ship                          &63.47  &64.64  &63.48  &\textbf{65.96}  \\
expressway toll station       &63.01  &\textbf{65.71}  &61.63  &64.82  \\
dam                           &61.61  &68.70  &64.54  &\textbf{71.31}  \\
harbor                        &60.05  &60.40  &60.47  &\textbf{62.03}  \\
bridge                        &30.48  &36.74  &34.24  &\textbf{37.94}  \\
vehicle                       &42.60  &47.63  &43.12  &\textbf{49.66}  \\
windmill                      &35.32  &41.99  &40.76  &\textbf{46.72}  \\  \midrule
average                          &62.44  &65.13  &64.12  &\textbf{66.46} \\ \bottomrule
\end{tabular}
} \label{tab:rrsisd_class}
\end{table}

\begin{figure*}
  \centering
  \includegraphics[width=\linewidth]{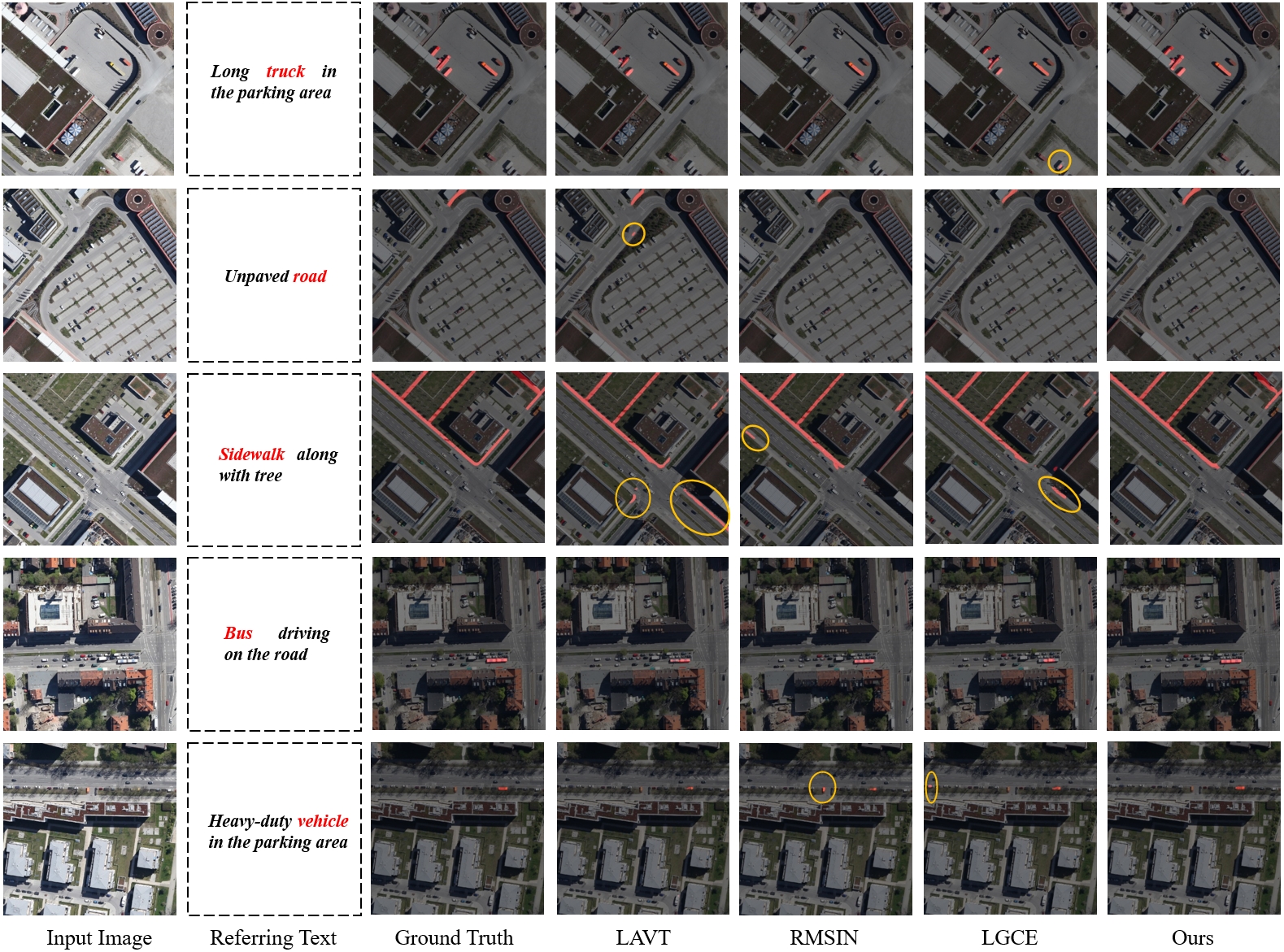}\\
  \caption{ Qualitative comparisons of different methods on RefSegRS dataset. The predicted masks are superposed on the original images and false alarms are circled in yellow. (Best view in Zoom)   
  }\label{pic:cmp_RefSegRS}
\end{figure*}

\begin{figure*}
  \centering
  \includegraphics[width=\linewidth]{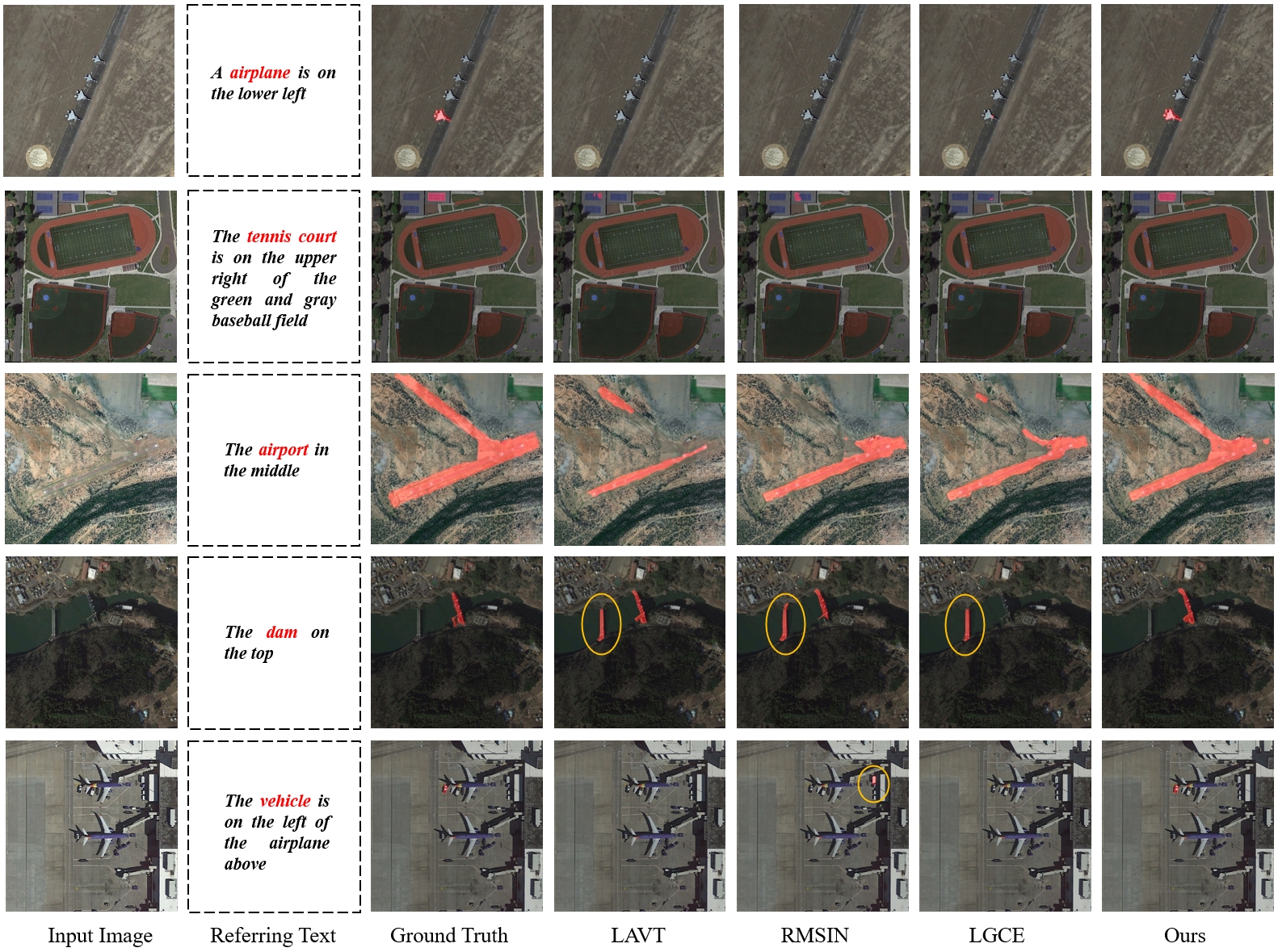}\\
  \caption{ Qualitative comparisons of different methods on RRSIS-D dataset. The predicted masks are superposed on the original images and false alarms are circled in yellow. (Best view in Zoom)   
  }\label{pic:cmp_RRSIS}
\end{figure*}

\subsection{Dataset and Metrics}

In the paper, we use two public remote sensing datasets, RefSegRS \cite{yuan2024rrsis} and RRSIS-D \cite{liu2024rotated}, to evaluate the effectiveness of the proposed method. These datasets were recently introduced, contributing to the advancement of the RRSIS task.

\begin{itemize}
\item RefSegRS \cite{yuan2024rrsis}. This dataset contains 4,420 image-text-label triplets in total. The training set has 2,172 triplets, the validation set has 431 triplets, and the rest 1,817 triplets are in the test set. The whole dataset covers 14 categories including road, vehicle, car, van, buliding and etc, with five attribute tags used to describe these ground objects. The image size is $512 \times 512$ and the spatial resolution is 0.13m. 

\item RRSIS-D \cite{liu2024rotated}. Compared with the RefSegRS, the RRSIS-D is a larger benchmark and comprises a collection of 17,402 images, masks, and referring expressions, with 12,181 for training, 1,740 for validation, and the rest 3,481 for testing. RRSIS-D contains 20 categories for the semantic labels and referring expressions, such as airplane, golf field, expressway service area, baseball field, stadium, and etc. The image size in this dataset is $800 \times 800$ with spatial resolutions ranging from 0.5m to 30m. 

\end{itemize}

Following some earlier works \cite{yang2022lavt, yuan2024rrsis, liu2024rotated}, we employ overall Intersection-over-Union (oIoU) and mean Intersection-over-Union (mIoU) to evaluate the overall results of different methods. Specifically, oIoU computes the ratio of the total intersection area to the total union area across the entire test set, thereby giving greater weight to large ground objects. 
The mIoU represents the average IoU computed between the predictions and their corresponding ground truths across all test samples, which treats large and small ground objects equally.
Moreover, precisions at threshold values of 0.5 to 0.9 (denoted as Pr@X) are also utilized to measure the ratio of test images that pass a specific IoU threshold.

\subsection{Comparisons with Other Methods}

We compare the proposed method with some state-of-the-art for referring image segmentation on the RefSegRS and RRSIS-D datasets. Among these methods, LGCE \cite{yuan2024rrsis} and RMISN \cite{liu2024rotated} are specifically designed for remote sensing images, and the others are for natural images. The results of different methods are provided in Table \ref{tab:refsegrs_overall} through Table \ref{tab:rrsisd_class}. For a fair comparison, we reimplement some state-of-the-art including LAVT \cite{yang2022lavt}, CrossVLT \cite{cho2023cross}, LGCE \cite{yuan2024rrsis}, and RMSIN \cite{liu2024rotated}, where the total number of train epochs for RefSegRS is set to 60 and the one for RRSIS-D is 40. Meanwhile, for some early published approaches, we take these results reported in LGCE \cite{yuan2024rrsis} and RMISN \cite{liu2024rotated}.

\textit{1) Quantitative Results on RefSegRS Dataset.} Table \ref{tab:refsegrs_overall} carefully lists the overall results of different methods on the RefSegRS. It can be seen that our proposed method outperforms other methods across all the metrics on this dataset. Particularly, our method obtains gains of 6.77\% in mIoU over the second-best LGCE. 
To further demonstrate the effectiveness of our method, we provide detailed comparisons of the fine-grained categories. The RefSegRS contains 14 kinds of scenes and the referring segmentation results of mean IoU for different categories are shown in Table \ref{tab:refsegrs_class}.
The results clearly show that the performance of referring segmentation varies significantly across different ground objects. For instance, ``road marking'' proves challenging to segment, while ``impervious surface'' is comparatively easier to recognize. In most categories, our method achieves higher mIoU values than LGCE, RMSIN, and LAVT. Furthermore, the average mIoU of our proposed method is substantially higher than that of the other three methods, demonstrating its effectiveness in handling diverse ground objects.

\textit{2) Quantitative Results on RRSIS-D Dataset.} Compared to the RefSegRS Dataset, RRSIS-D is a larger dataset with 20 categories of ground objects, providing training samples to optimize the models. The overall results are presented in Table \ref{tab:rrsisd_overall}. Likewise, our method achieves the best performance on this dataset in terms of mIoU, oIoU, and from Pr@0.5 to Pr@0.8. Specifically, the proposed method obtains gains of 1.74\% in mIoU over the second-best RMSIN.
We have also calculated the segmentation results for each category, as presented in Table \ref{tab:rrsisd_class}. Compared to RefSegRS, the ground objects in the RRSIS-D dataset are more challenging to identify due to their diverse and varying scales.
Our method obtains the best performance on most ground objects, including road, vehicle, car, van and etc., and achieves the highest average mIoU with 1.33\% higher than the second-best RMSIN.

\textit{3) Qualitative Comparisons.} We here provide some qualitative comparisons with LAVT, RMSIN, and LGCE on these two datasets. Fig. \ref{pic:cmp_RefSegRS} shows several segmentation results referring to the corresponding texts of the RefSegRS dataset, including truck, road, sidewalk, bus, and vehicle scenes which are marked in red. Moreover, Fig. \ref{pic:cmp_RRSIS} illustrates the outcomes of RRSIS-D dataset covering several ground objects such as airplane, tennis court, airport, dam and vehicle. Some false alarms of different methods are circled in yellow.
Additionally, it achieves more precise localization of ground objects while reducing false alarms. These visual comparisons highlight the robustness of the proposed method across diverse ground objects and scales, ranging from tiny vehicles to medium-sized dams and large airports.


\subsection{Ablation Studies}\label{section:seg}

We conduct a series of ablation experiments on the test subset of RefSegRS dataset to validate the effectiveness of core components of our method. 

\begin{figure}
  \centering
  \includegraphics[width=\linewidth]{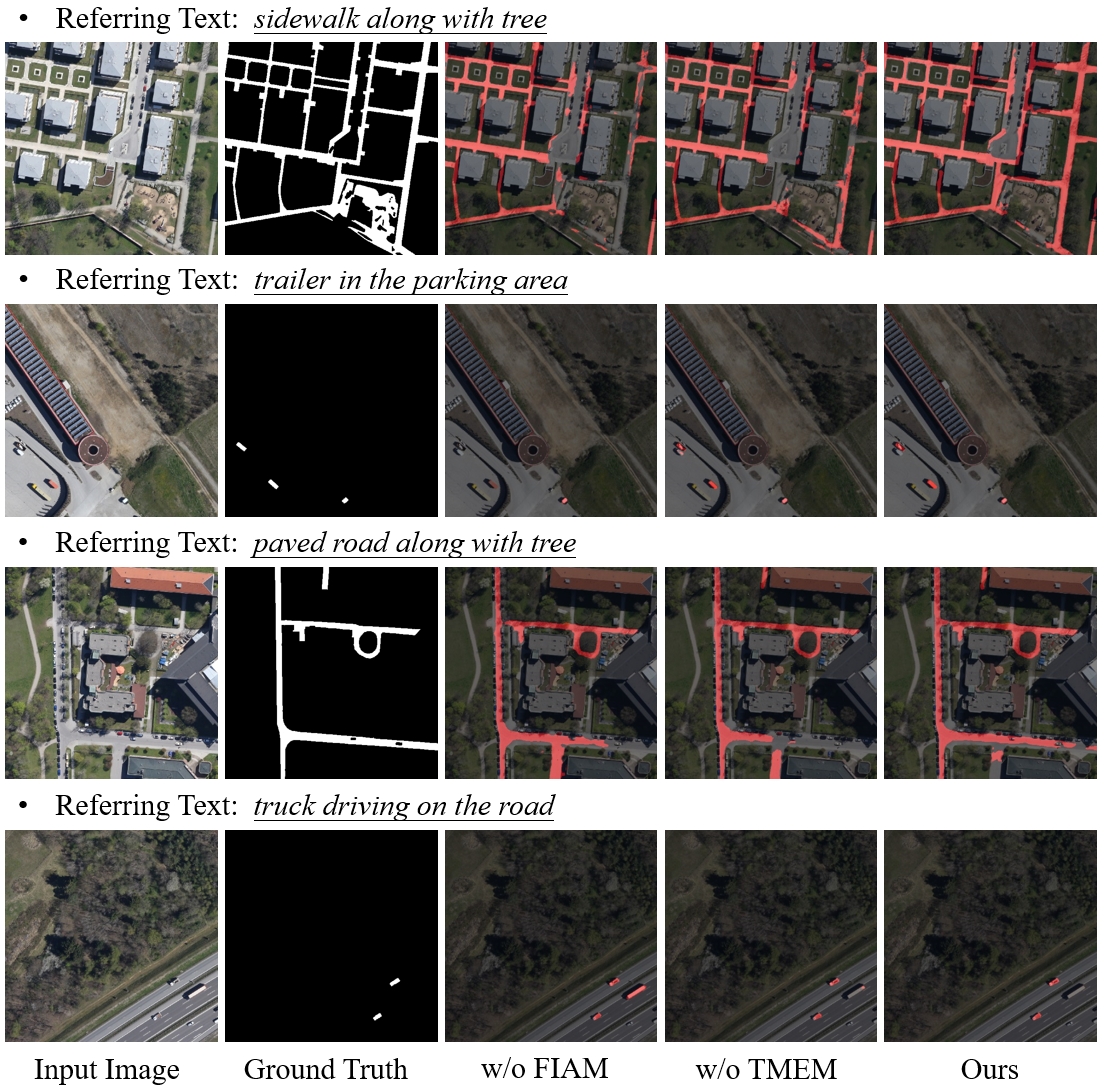}\\
  \caption{  Qualitative comparisons of different settings on RefSegRS dataset. (Best view in Zoom)   
  }\label{pic:ablation-vis}
\end{figure}

\textit{1) Effectiveness of FIAM and TMEM.} We design some experiments to assess the importance of FIAM and TMEM, and the results are listed in Table \ref{tab:components}. The baseline without FIAM and TMEM leverages a traditional image-text alignment used in LAVT. As shown in Table \ref{tab:components}, the introduction of FIAM can largely improve the segmentation results, where mIoU obtains an increase of 4.44\%. The combination of the FIAM and TMEM further promotes the performance of the proposed method.
To demonstrate the effectiveness of these two modules, we visually compare the segmentation maps on some samples of the RefSegRS dataset, as shown in Fig. \ref{pic:ablation-vis}. The results indicate that the proposed method, incorporating FIAM and TMEM, achieves superior performance and improved segmentation outcomes.

\begin{table}
  \centering
   \caption{Ablation studies on the fine-grained image-text alignment module (FIAM) and Text-aware Multi-scale Enhancement Module (TMEM).}
  \resizebox{\linewidth}{!}{
  \begin{tabular}{cc|ccccc}
    \toprule
    FIAM & TMEM & P@0.5 & P@0.7 & P@0.9 & oIoU & mIoU \\
    \midrule
          &   & 78.37  & 43.04  & 2.86  & 74.90  & 62.24   \\
     \checkmark  &   & 83.21 & 57.29  & 4.79  & 77.83  & 66.68\\
      & \checkmark & 80.96  & 53.99   & 4.29  & 75.62  & 65.39 \\     
     \rowcolor{gray!25} \checkmark & \checkmark & 84.09 & 61.86  & 7.10  & 78.32 & 68.67 \\
    \bottomrule
  \end{tabular}
  }
  \label{tab:components}
\end{table}

2) \textit{Effect of different designs of FIAM.} In order to provide an in-depth understanding of FIAM, we carry out some experiments to explore the effect of different designs of FIAM. For this point, we remove some key components of this module to record the change in metrics. As provided in Table \ref{tab:FIAM}, we explore the influences of channel modulation (C.M.), ground object branch (G.O.B), and spatial positional branch (S.P.B). It is obvious that through the integration of these components, the proposed method obtains better performance. Furthermore, this verifies the effectiveness of the fine-grain image-text alignment. 

\begin{table}
  \centering
   \caption{Ablation on Effect of different designs of FIAM.}
  \resizebox{\linewidth}{!}{
  \begin{tabular}{ccc|ccccc}
    \toprule
     C.M. & G.O.B. &S.P.B. & P@0.5 & P@0.7 & P@0.9 & oIoU & mIoU \\
    \midrule
                 &     &   &80.96  &53.99 &4.29 &75.62 &65.39 \\ 
     \checkmark  &     &  &81.40  &53.99 &4.73 &76.07 & 65.76     \\
     \checkmark  &\checkmark  &  & 83.27  & 57.46  &5.34    &77.40  &67.01     \\
     \rowcolor{gray!25} \checkmark & \checkmark &\checkmark  & 84.09 & 61.86  & 7.10  & 78.32 & 68.67 \\     
    \bottomrule
  \end{tabular}
  }
 
  \label{tab:FIAM}
\end{table}

\begin{figure}
  \centering
  \includegraphics[width=8cm]{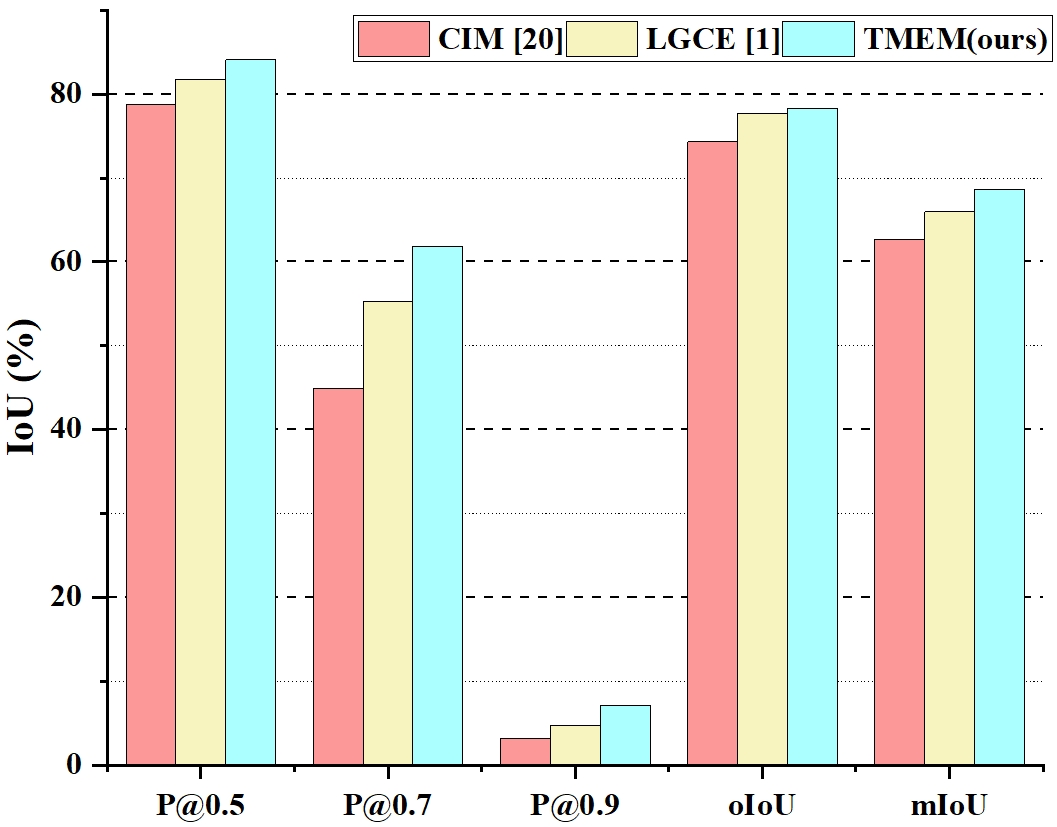}\\
  \caption{ The comparisons of different designs of multi-scale fusion.  
  }\label{pic:TMEM_abl}
\end{figure}

3) \textit{Effect of different designs of multi-scale fusion.} To further demonstrate the efficacy of the proposed TMEM, we here use the other two designs of multi-scale fusion to be comparisons, i.e., Cross Intersection Module (CIM) \cite{liu2024rotated} and Language-Guided Cross-scale Enhancement (LGCE) \cite{yuan2024rrsis}. We use the CIM or LGCE to replace the TMEM and the other designs remain the same. 
Fig. \ref{pic:TMEM_abl} shows that the proposed TMEM outperforms CIM and LGCE across all metrics, highlighting the importance of referring text in multi-scale feature enhancement and demonstrating the effectiveness of the proposed TMEM.

\section{Conclusions}\label{sec:conclusion}
In this paper, we propose a new referring image segmentation method for remote sensing, named FIANet, from the perspective of fine-grained image-text alignment. Specifically, we design a Fine-grained Image-text Alignment Module (FIAM) to exploit the subtle association between the visual and linguistic features and learn better discriminative multi-modal representations.
Moreover, to handle the various scales of ground objects in remote sensing, we introduce a Text-aware Multi-scale Enhancement Module (TMEM) to adaptively perform cross-scale fusion and intersections under text guidance. 
We evaluate the effectiveness of the proposed methods on two public referring remote sensing datasets including RefSegRS and RRSIS-D, demonstrating that our method achieves superior performance over several state-of-the-art methods. Meanwhile, comprehensive ablation experiments also verify the effectiveness of FIAM and TMEM.

While the proposed method achieves promising results in referring remote sensing image segmentation, there remains significant room for further exploration in this task. Future work could focus on developing more efficient multi-modal fusion strategies between image and linguistic features to enhance applicability in practical scenarios. Additionally, foundation models have demonstrated great potential in computer vision and remote sensing tasks, making their integration into this task a valuable direction for future research.

{
\bibliographystyle{IEEEtran}
\bibliography{refbib}
}

\end{document}